\documentclass{article}

\usepackage{graphicx}
\usepackage[preprint]{nips}

\usepackage[utf8]{inputenc} 
\usepackage[T1]{fontenc}    
\usepackage{hyperref}       
\usepackage{url}            
\usepackage{booktabs}       
\usepackage{amsfonts}       
\usepackage{nicefrac}       
\usepackage{microtype}      
\usepackage{xcolor}         
\usepackage{amsmath}
\usepackage{multirow}
\usepackage{booktabs}
\usepackage{bbding}
\usepackage{bm}

\makeatletter
\newcommand{\biggg}{\bBigg@{1.1}}

\makeatother

\title{Advancements and Trends in Ultra-High-Resolution Image Processing: An Overview}

\author{Zhuoran Zheng$^{1}$~~~Boxue Xiao$^{2}$\\
	$^{1}$CSE, Nanjing University of Science and Technology~~~~$^{2}$Hong Kong Polytechnic University\\
	{\tt\small zhengzr@njust.edu.cn}
}

\begin{document}
	\bibliographystyle{plain}
	\maketitle

\begin{abstract}
Currently, to further improve visual enjoyment, Ultra-High-Definition (UHD) images are catching wide attention.
Here, UHD images are usually referred to as having a resolution greater than or equal to $3840 \times 2160$.
However, since the imaging equipment is subject to environmental noise or equipment jitter, UHD images are prone to contrast degradation, blurring, low dynamic range, etc.
To address these issues, a large number of algorithms for UHD image enhancement have been proposed.
In this paper, we introduce the current state of UHD image enhancement from two perspectives, one is the application field and the other is the technology.
In addition, we briefly explore its trends.
\end{abstract}

\section{Application Field}
Here, we introduce several algorithms for implementing image enhancement on UHD image degradation scenarios. Notably, most of these methods are implemented to run on a single GPU shader (RTX3090 or V100) in real-time ($>$30fps).

\noindent \textbf{Image dehazing.}
UHD image dehazing started with~\cite{zheng2021ultra}, which employs a bilateral learning approach to enhance 4K images. After that, Xiao et al.~\cite{xiao2024single} propose a pyramid model to enhance the dehazing effect of 4K images. Recently, Zhuang et al.~\cite{10323236} proposed a novel Mixer model to conduct real-time dehazing of 4K images. In addition to the above methods, AODNet~\cite{li2017all} can also achieve the dehazing of 4K images in real-time. It is worth noting that the way of dividing the patches is not suitable for the image-dehazing task, it can suffer from the grid effect~\cite{zheng2021ultra}.

\noindent \textbf{HDR enhancement.}
HDRNet~\cite{gharbi2017deep} is a very well-known network that enhances a UHD degraded image in real-time.
HDRNet introduces bilateral filters to deep learning to reconstruct degraded images of arbitrary resolution.
Later, Zheng et al.~\cite{zheng2021ultra1} improved the algorithm to make it more efficient in reconstructing a low dynamic range UHD image.
Currently, a large number of algorithms~\cite{yao2023bidirectional,zhang2023multi} are focused on solving the problem of UHD HDR video.
In addition, exposure correction on UHD images is also a concern~\cite{zhou20234k,wang2021real,liu20234d,zeng2020learning,yang2022adaint,zhang2022dualbln,yang2022seplut,jiang2023meflut,wang2023lgabl}.

\noindent \textbf{Low light image enhancement.}
The enhancement of UHD low-light images begins with~\cite{lin2022uhd}, which employs a bilateral learning approach to enhance a dim image.
Subsequently, a large number of methods and benchmarks have been proposed for UHD low-light image enhancement.
For example, wang et al.~\cite{wang2023ultra} propose a Transformer-based approach to spatial domains.
Li et al.~\cite{li2023embedding} propose to solve the problem of contrast degradation of UHD images in the frequency domain.

\noindent \textbf{Underwater image enhancement.}
Wei et al.~\cite{wei2022uhd} propose a two-domain learning method to enhance the visibility of underwater images. Jiang et al.~\cite{jiang2023five} propose a lightweight network that can run a UHD water image on a single GPU.

\noindent \textbf{Image super-resolution.}
Zhang et al.~\cite{zhang2021benchmarking} first propose a benchmark for 4K image super-resolution.
Currently, this task has seen the emergence of a professional competition to evaluate the performance of 4K image super-resolution algorithms~\cite{zamfir2023towards,conde2023efficient,gankhuyag2023lightweight}.

\noindent \textbf{Denoising and demosaicing.}
Guan et al.~\cite{guan2022memory} propose a convolution with a memory architecture to achieve the denoising of UHD images.
Yu et al.~\cite{yu2022towards} have accomplished the demosaicing of 4K images.

\noindent \textbf{Image deblurring.}
Zheng et al.~\cite{zheng2022uhd} propose for the first time an algorithm for deblurring UHD images in real-time.
In addition, there is an algorithm~\cite{deng2021multi} for deblurring UHD videos.

\noindent \textbf{Others.}
4K image matting~\cite{sun2023ultrahigh}, inpainting~\cite{schrader2023efficient}, UDC~\cite{luo2022under,zhengimage} is also starting to come into its own.

\section{Method}
\begin{figure*}[h]
	\centering%
	\includegraphics[width=0.945\textwidth]{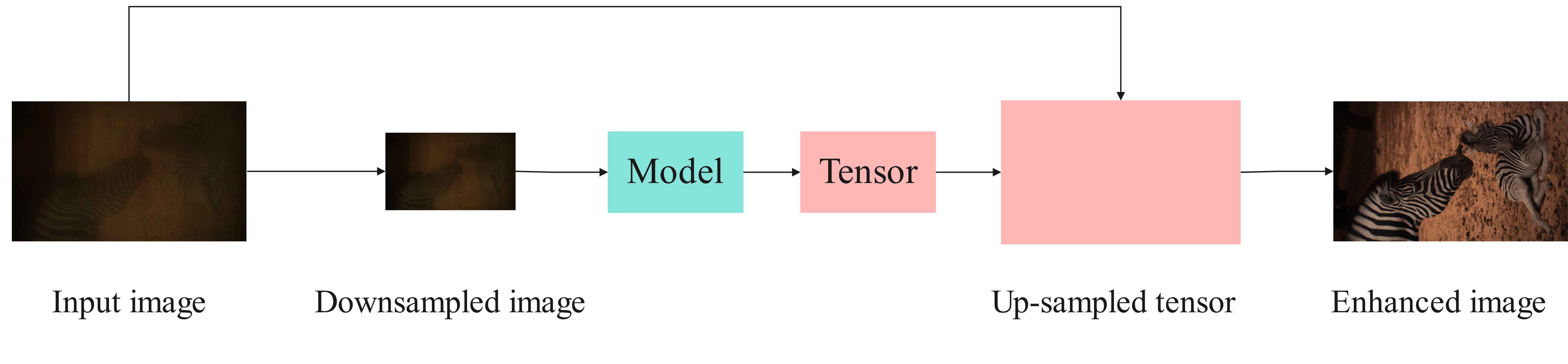}
	\caption{\textbf{A flow diagram of popular UHD image enhancement algorithms}. It is worth noting that some tensors may be splashed in using human a priori knowledge, such as the LUT model. Some tensors are progressively upsampled such as the pyramid model. In general, none of the existing methods can avoid the paradigm where the input image is downsampled.}
	\label{fig:fig1}
	\vspace{-0mm}
\end{figure*}

Up to now, the processing methods for UHD images are mainly categorized into 3 groups, bilateral learning~\cite{zheng2021ultra,gharbi2017deep,zheng2021ultra1,wang2023lgabl,lin2022uhd,xu2023direction,xia2020joint,xu2021bilateral,zhou20234k}, LUT~\cite{yang2022seplut,jiang2023meflut,zhang2022dualbln,yang2022adaint,liu20234d,wang2021real,li2022mulut,jo2021practical}, and pyramid~\cite{xiao2024single,jiang2023five,liang2021high}.
As shown in Figure~\ref{fig:fig1}, overall, all of the above methods require downsampling the UHD image and feeding it into a model of higher complexity to learn an attention-like tensor.
Finally, this tensor is upsampled (bilinear interpolation or subpixel convolution) to the same resolution as the original image acting on the original input signal.
It is worth noting that the model can run in real-time thanks to the low flops rather than the optimization of the operators.
%

%
%
%
%
%

%
%
%
%

\vspace{-0mm}

\section{Discussion and Future Work}
So far, tasks such as image de-rain, image de-reflection, and image de-dropping have not been addressed.
This may be due to the difficulty in constructing the dataset.
In this paper, we note that there are 3 hard to overcome problems with existing UHD image enhancement tasks.
1) The training process of pairs of UHD images inevitably requires downsampling of the images, which can result in the trained model receiving ``knowledge'' that may differ from the real 4K images.
2) For now, common convolutional and MLP operators are not tailor-made to address UHD image enhancement efficiently.
3) UHD synthetic datasets are expensive to build, for example, dehazing 10,000 pairs of 4K images takes 1 week of rendering on a single GPU.

In the future, one of the first issues we will address is to change the learning paradigm shown in Figure~\ref{fig:fig1}.
This is due to the inevitable loss of information that comes with downsampled images.
Second, facing this single hyperscale example, we may need to refer to some fine-tuning scheme of the larger model to alleviate this problem.
Predictably, UHD images are a worthwhile direction to think about in the field of image enhancement.
%

%

\bibliography{4KSR}

\end{document}